\documentclass[conference]{IEEEtran}
\IEEEoverridecommandlockouts
\usepackage{cite}
\usepackage{amsmath,amssymb,amsfonts}
\usepackage{algorithmic}
\usepackage{graphicx}
\usepackage{textcomp}
\usepackage{xcolor}
\usepackage{booktabs}
\usepackage{multirow}
\def\BibTeX{{\rm B\kern-.05em{\sc i\kern-.025em b}\kern-.08em
    T\kern-.1667em\lower.7ex\hbox{E}\kern-.125emX}}
\begin{document}

\title{Improving Real-Time Omnidirectional 3D Multi-Person Human Pose Estimation with People Matching and Unsupervised 2D-3D Lifting\\
}

\author{\IEEEauthorblockN{1\textsuperscript{\textsection} Pawel Knap}
\IEEEauthorblockA{\textit{Vision Learning and Control, (ECS)} \\
\textit{University of Southampton}\\
Southampton, UK \\
pmk1g20@soton.ac.uk}
\and
\IEEEauthorblockN{1\textsuperscript{\textsection} Peter Hardy}
\IEEEauthorblockA{\textit{Vision Learning and Control, (ECS)} \\
\textit{University of Southampton}\\
Southampton, UK \\
p.t.d.hardy@soton.ac.uk}
\and
\IEEEauthorblockN{2\textsuperscript{nd} Alberto Tamajo}
\IEEEauthorblockA{\textit{Vision Learning and Control, (ECS)} \\
\textit{University of Southampton}\\
Southampton, UK \\
at2n19@soton.ac.uk}
\and
\IEEEauthorblockN{3\textsuperscript{rd} Hwasup Lim}
\IEEEauthorblockA{
\textit{Korean Institute of Science and Technology}\\
Seoul, South Korea \\
hslim@kist.re.kr}
\and
\IEEEauthorblockN{4\textsuperscript{th} Hansung Kim}
\IEEEauthorblockA{\textit{Vision Learning and Control, (ECS)} \\
\textit{University of Southampton}\\
Southampton, UK \\
h.kim@soton.ac.uk}
}

\maketitle
\begingroup\renewcommand\thefootnote{\textsection}
\footnotetext{Both authors contributed equally to this research. \\ This work was supported by the Korea Institute of Science and Technology (KIST) Institutional Program (Project No. 2E32303)}
\endgroup
\begin{abstract}
Current human pose estimation systems focus on retrieving an accurate 3D global estimate of a single person. Therefore, this paper presents one of the first 3D multi-person human pose estimation systems that is able to work in real-time and is also able to handle basic forms of occlusion. First, we adjust an off-the-shelf 2D detector and an unsupervised 2D-3D lifting model for use with a 360$^\circ$ panoramic camera and mmWave radar sensors. We then introduce several contributions, including camera and radar calibrations, and the improved matching of people within the image and radar space. The system addresses both the depth and scale ambiguity problems by employing a lightweight 2D-3D pose lifting algorithm that is able to work in real-time while exhibiting accurate performance in both indoor and outdoor environments which offers both an affordable and scalable solution. Notably, our system's time complexity remains nearly constant irrespective of the number of detected individuals, achieving a frame rate of approximately 7-8 fps on a laptop with a commercial-grade GPU.
\end{abstract}


\section{Introduction}
\begin{figure}[h!]
    \centering
    \includegraphics[width=0.34\textwidth]{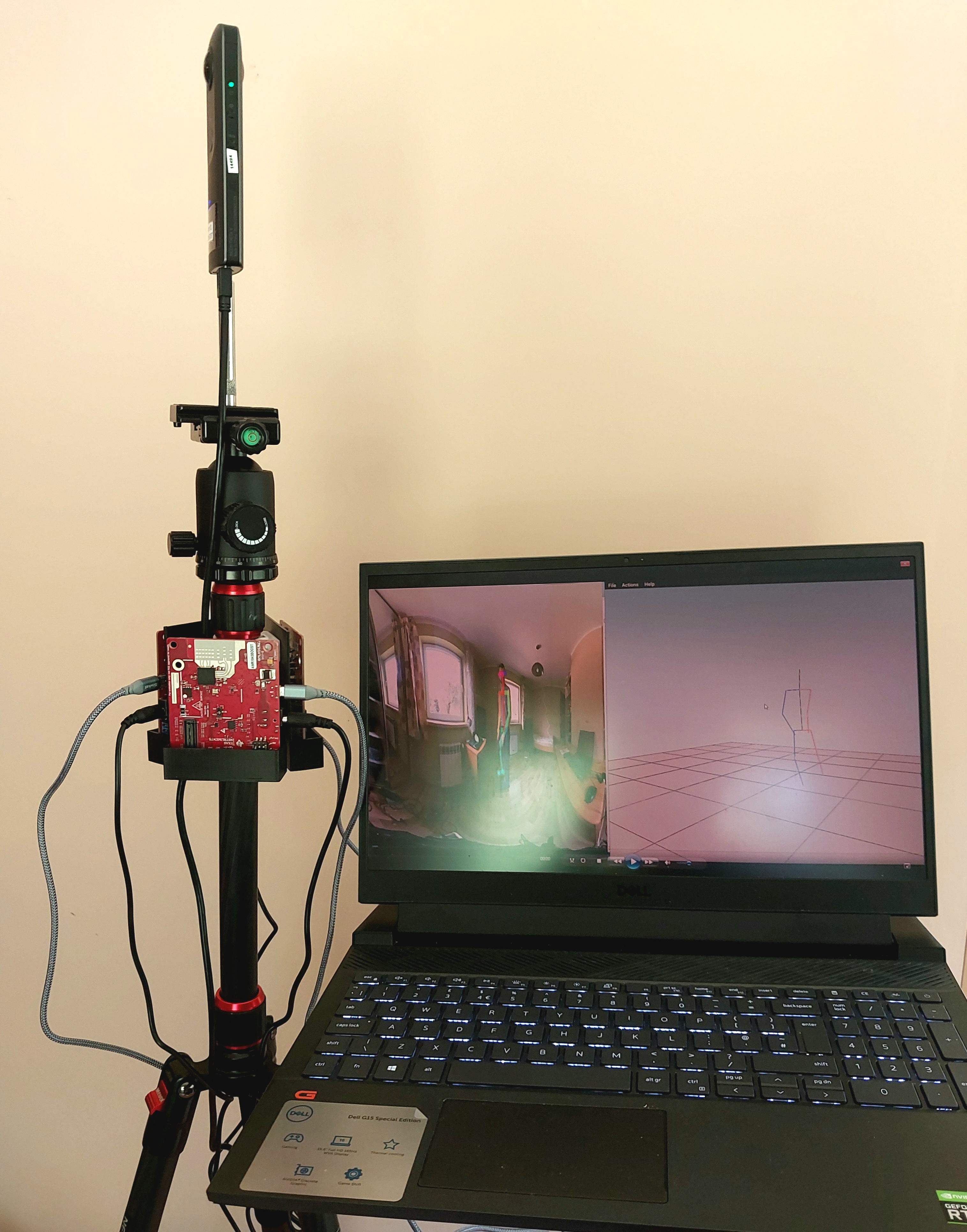}
    \caption{Our experimental setup consists of a laptop with RTX 3060, the Ricoh Theta V omindirectional camera and three TI AWR1843BOOST mmWava radars.}
    \label{setup}
\end{figure}
3D human pose estimation (HPE) from a single camera is an important task, with various applications such as security, 3D animation and physical therapy \cite{1,2,3}.Nevertheless, obtaining precise 3D global coordinates from a single viewpoint remains challenging due to inherent perspective ambiguities. Previous strategies have often resorted to merging RGB cameras with laser or infrared (IR) depth sensors \cite{6,7}. While effective, these methods are marred by cost issues associated with laser-based sensors and sub-optimal performance of IR-based sensors in outdoor environments due to sunlight, thus limiting their widespread adoption. To address this, radar-based methods have emerged as a cost-effective solution for both indoor and outdoor scenarios. To this end we present one of a new approach that utilises mmWave radars and omnidirectional cameras, allowing us to accurately reconstruct multiple people within a $360^\circ$ scene. To detect and lift the pose to 3D its own local coordinate space we utilise two off-the-shelf methods. The first is the 2D detector OpenPose \cite{14} which we use to obtain the 2D keypoint locations of people within the image space. The second is our preliminary 2D-3D lifting network LInKs \cite{hardy2023links}, an unsupervised 2D-3D lifting approach that is also able to handle basic forms of occlusion. This is especially important due to occlusion being highly prominent in omnidirectional video, be that self-occlusion or occlusion from an object. We further refine our methodology based on prior work \cite{last}, introducing an alternative approach to match individuals in both the image and radar domains. This enhancement not only elevates matching accuracy by 4.63\% but also reduces the absolute error in placing poses within the 3D coordinate system. Consequently, this paper represents a significant evolution from our previous work \cite{last}, encompassing essential advancements, including radar and camera calibration and an improved matching algorithm. These updates translate into substantial improvements in precision, accuracy, and our ability to effectively handle occlusions. The experimental setup of our approach is visually depicted in Fig. \ref{setup}.

\section{Methodology}
\begin{figure}[h!]
    \includegraphics[width=0.45\textwidth]{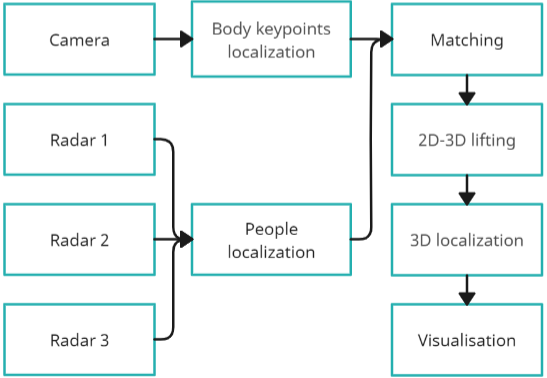}
    \caption{Overview of our approach. We use the video from an omnidirectional camera to obtain 2D body keypoints in the image space. Simultaneously we use 3 radar sensors to localise each person in our global 3D coordinate system. We then match these detected 2D poses to our radar's depth estimate. Next, these 2D poses are lifted into 3D and to finalise we transform their predicted 3D coordinates to be within our global coordinate system.}
    \label{picture: System pipeline}
\end{figure}

The core of our method resolved around transforming the 2D keypoints detected into the image space into 3D keypoints that are within our global coordinate system. An overview of each stage can be seen in Fig. \ref{picture: System pipeline}.
\subsection{Camera and Radar Calibration}
To accurately localise the people within our global coordinate system we calibrated both the camera and radars. The camera was calibrated using the method introduced in Zang \emph{et al.} \cite{zhang2000flexible}. To calibrate our radars, for each of our radars $(x, z)$ direction, an affine transformation was obtained using the Levenberg–Marquardt (LM) algorithm. To perform this multiple radar readings were collected by placing an individual at different known radar coordinates space 50cm apart. These readings were recorded for several seconds and the average value was then compared to the correct location to obtain our affine transformation.

\subsection{Data Fetching, 2D Keypoint Localisation and Person Matching}
The first stage of our proposed system involved acquiring data from each of the sensors, specifically an image frame from the camera and localisation data from each of the radars. As synchronisation is crucial to ensure consistency between the camera and radars, we separate the data obtained from each sensor into different threads. The camera data is obtained in the main thread, whereas the radar data is obtained via separate threads. The camera thread then signals the radar threads to add their data to a shared queue, ensuring synchronisation. To obtain the 2D keypoints, $\mathbf{x}$, of people in our image we used OpenPose \cite{14}, a popular 2D pose detector that is capable of detecting multiple people in real-time. To associate the 2D keypoints of people in the image space with their corresponding radar data, we employed a binary search tree method with a threshold value. The matching technique relied on the disparity between the average image $x$ coordinate of a person detected by OpenPose, denoted as $\bar{x} = \frac{\mathbf{x}}{N}$ where $N$ is the number of keypoints detected (15 in our study), and the radars coordinates transformed into the image coordinate space through a learned transform. This transform is described in the pseudo-inverse section of Oh \emph{et al.} \cite{oh}. Lastly, in our simultaneous stage of radar localisation, we used the people counting algorithm introduced by Garcia \cite{26} to acquire the $(x,z)$ coordinates of the people within our scene. These were then transformed into the common coordinate system to calculate the distance of each person from the camera.

\subsection{Unsupervised 2D-3D Lifting and 3D placement}
To lift the detected 2D pose to 3D we employed a recent 2D-3D lifting network known as LInKs \cite{hardy2023links}. We chose this due to its accuracy when generalising to unseen poses, as well as its ability to handle the most common forms of pose occlusion. Similar to other unsupervised 2D-3D lifting networks, the LInKs algorithm does not predict the absolute depth of each keypoint, but instead the depth off-set ($\hat{d}$) of each keypoint relative to a root joint (typically the pelvis), when the pose is assumed to be $c$ units from the camera. The final 3D location of a specific keypoint, $\mathbf{x}_i$, was then obtained via perspective projection:
\begin{equation}
\begin{split}
\mathbf{x}_i &= (x_i\hat{z}_i, y_i\hat{z}_i, \hat{z}_i), \\
\mathbf{where } \quad \hat{z}_i &= \max(1, \hat{d}_i + c).
\end{split}
\end{equation}
where $d_i$ was our models' depth-offset prediction for keypoint $i$. As LInKs was originally trained on Human3.6M \cite{h36m_pami}, which uses different keypoints than those detected by OpenPose, we retrained it on the OpenPose keypoints present in the Human3.6M dataset. Once we had lifted our 2D pose to 3D, we now had to transform it from its local coordinate system, where the root joint was at position (0,0,$c$), to our global coordinate system. To do this we subtracted $c$ from the pose and added the $x$ and $z$ coordinates from our radar sensors. Additionally, to maintain ground-plane contact the $y$ coordinates of the pose were updated by subtracting the $y$ coordinate of the lowest ankle.
\begin{figure*}[t!]
    \centering
    \includegraphics[width=\textwidth]{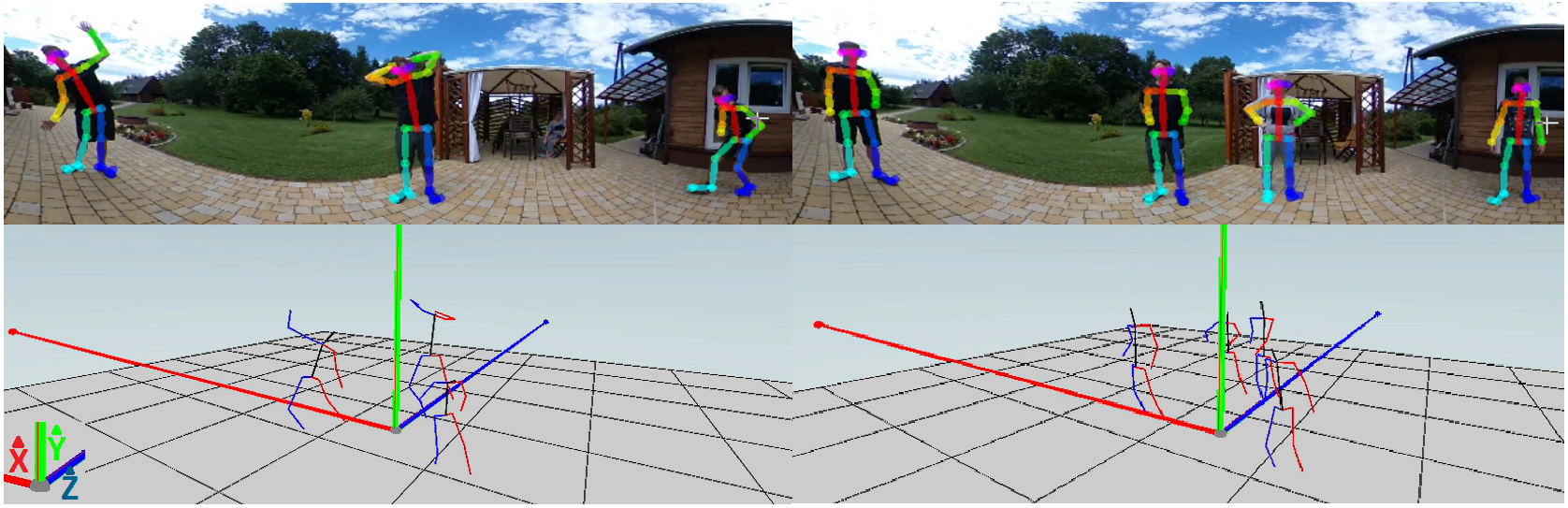}
    \caption{Qualtative results of our approach. The above images show the input frame to our model with poses captured by OpenPose. The bottom images show the corresponding reconstructed 3D poses in our global 3D coordinate system. All pictures are partially cropped around the top and bottom. }
    \label{qualitative_1}
\end{figure*}
\section{Results}
Here we present the quantitative of our improved people matching algorithm, 2D-3D human pose lifting model and the radar localisation error for the $x$ and $z$ coordinate. The qualitative results of our approach can be seen in Fig. \ref{qualitative_1}. In our experiment we used a commercial RTX 3060, Ricoh Theta V omnidirectional camera and three TI AWR1843BOOST mmWave radars. However, it is worth noting that it will work with any GPU, mmWave radars or omnidirectional camera as long as the latter outputs video frames in the equirectangular format. 

\subsection{Results of matching people in the image and radar}
In our preliminary work, we matched people by using the angle between people detected in the radar and camera relative to the camera's $x$ coordinate. In our approach, we implemented an improved method that solely focuses on the $x$ coordinate in the camera space which illustrated a significant improvement. To demonstrate this we calculated the matching error as a \% which represents the absolute difference between the radar and camera values of an individual, divided by the camera values as seen in Table. Our results for this can be seen in Table \ref{tab:matching_results}
\begin{table}[h]
\caption{Showing the matching error of our preliminary work and our improved approach. The value is a \% obtained by computing the absolute difference between the corresponding radar and camera values}
\centering
\resizebox{\columnwidth}{!}{%
\begin{tabular}{@{}lccc@{}}
\toprule
                 & Radar 1 $\downarrow$    & Radar 2  $\downarrow$    & Radar 3  $\downarrow$                         \\ \midrule
Preliminary Work \cite{last} & 23.89\%  $\pm$ 6.57\% & 33.57\%  $\pm$ 50.55 & \multicolumn{1}{c}{66.89\%  $\pm$ 263.89} \\
Ours             & 2.52\% $\pm$ 2.51  & 9.44\%  $\pm$ 13.27  & 1.94\%  $\pm$ 1.52                        \\ \bottomrule
\end{tabular}%
}
\label{tab:matching_results}
\end{table}
\subsection{2D-3D Lifting and Occlusion Handling Results}
As previously mentioned we used the LInKs lifting network \cite{hardy2023links} for 2D-3D pose lifting. We trained LInKs unsupervised on Human3.6M \cite{h36m_pami} with identical training and model parameters to its original publication, with the only modification being adapting it to use the keypoints detected via OpenPose. We report the mean per joint position error which is the euclidean distance in millimetres between the ground truth keypoints and those within our reconstructed pose. For this we report both the error once our pose has been scaled to the ground truth (N-MPJPE) and once our pose was rigidly aligned to the ground truth. We also show the results in various occlusion scenarios. These can be seen in Table \ref{tab:2d-3dliftingtable}. As shown we achieved similar results in N-MPJPE under scenarios of no occlusion with a slightly higher PA-MPJPE. We attribute this to the OpenPose keypoints not using the spine and head-top keypoint which are relatively easy to estimate the depth-offset of from the pelvis, being that they are typically directly above in the majority of standing scenarios. This is why we see a similar N-MPJPE as our scaled poses have a very similar accuracy however a slightly higher PA-MPJPE as two relatively easy keypoints are not included in the alignment. 
\begin{table}[]
\caption{Showing the 2D-3D lifting results of using the LInKs model trained on the OpenPose keypoints of Human3.6M}
\centering
\resizebox{\columnwidth}{!}{%
\begin{tabular}{@{}llcc@{}}
\toprule
Method            & Occlusion        & PA-MPJPE & N-MPJPE \\ \midrule
LInKs \cite{hardy2023links}           & None             & 33.8     & 61.6    \\
Ours (Recreation) & None             & 37.2     & 61.7    \\
Ours (Recreation) & Left Arm         & 52.1     & 78.1    \\
Ours (Recreation) & Left Leg         & 46.0     & 73.2    \\
Ours (Recreation) & Right Arm        & 49.8     & 75.7    \\
Ours (Recreation) & Right Leg        & 44.5     & 71.6    \\
Ours (Recreation) & Left Arm \& Leg  & 62.0     & 86.0    \\
Ours (Recreation) & Right Arm \& Leg & 60.2     & 83.7    \\
Ours (Recreation) & Both Legs        & 69.3     & 99.8    \\
Ours (Recreation) & Torso            & 88.4     & 122.0   \\ \bottomrule
\end{tabular}%
}
\label{tab:2d-3dliftingtable}
\end{table}
\begin{figure*}[h]
\centering
  \includegraphics[width=0.44\textwidth]{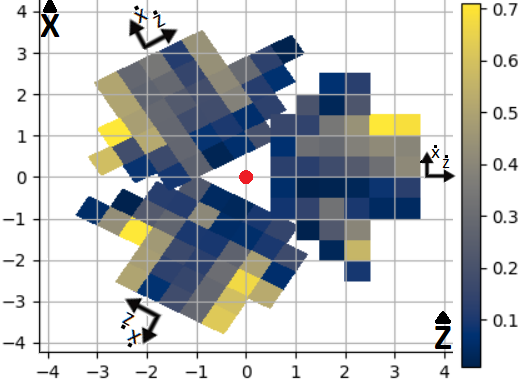}
  \hspace{1cm}
  \includegraphics[width=0.44\textwidth]{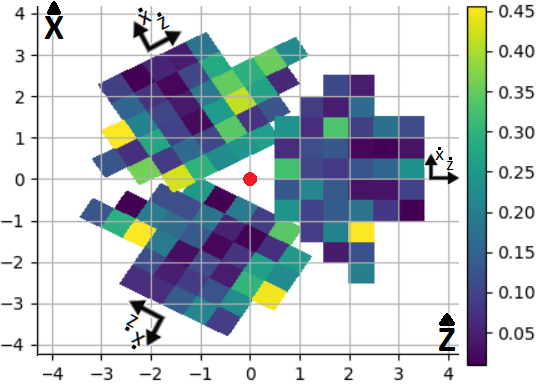}
  \label{fig:radar_localisation}
\caption{Showing the localisation error in metres at various points around our setup. The errors were evaluated in each radar's $\hat{x}$ (left) and $\hat{z}$ (right) directions. The figures represent these errors in the $(\mathbf{\hat{X}},\mathbf{\hat{Z}})$ 2D global coordinate system. The red dot marks the system location.}
\end{figure*}
\subsection{Radar Localisation Results}
To demonstrate the improvement in radar localisation due to our affine transformation we present the average absolute error in centimetres in the $x$ and $z$ direction for people within our scene. These results are presented in Table \ref{tab:localisation_error}, additionally Fig. \ref{fig:radar_localisation} visualises the localisation errors of objects from the radars in metres at various points around the system. As shown in our results our affine transformation has led to a reduction in the error along the $x$ and $z$ direction for nearly all radars while performing similarly for the $z$ direction for radar 1. Despite this, we note that there are still some errors present, especially when the subject is positioned at a $60^\circ$ angle from the centre of the radar despite the radars $120^\circ$ coverage. In addition, we noticed void spaces exist in these areas where none of our radars were able to detect our subjects. One possible remediation for this would be the inclusion of additional radars directed at these areas of high error.
\begin{table}[t]
\caption{Table showing the mean absolute error in centimetres in the x and z direction for each of our radars in our preliminary work and our new approach with affine transformation.}
\centering
\resizebox{0.6\columnwidth}{!}{%
\begin{tabular}{@{}cccc@{}}
\toprule
\multicolumn{1}{l}{Radar} & Direction & Preliminary \cite{last}  & Ours           \\ \midrule
\multirow{2}{*}{1}        & $x$         & 20.65          & 16.45 \\
                          & $z$         & 11.41 & 11.45          \\ \midrule
\multirow{2}{*}{2}        & $x$         & 26.19          & 24.86 \\
                          & $z$         & 15.39          & 10.77 \\ \midrule
\multirow{2}{*}{3}        & $x$         & 16.88          & 15.94 \\
                          & $z$         & 13.83          & 13.46 \\ \bottomrule
\end{tabular}%
}
\label{tab:localisation_error}
\end{table}
\section{Conclusion}
In conclusion, our real-time 3D multi-person detection system significantly improves our preliminary work \cite{last}, offering simplicity, robustness, and scalability. Challenges remain in system speed, range, and occlusion handling. Future research will focus on improving occlusion handling, optimizing algorithm speed, and expanding the system's range. The inclusion of additional radar units and higher-resolution cameras can eliminate limitations related to detection gaps and bent knees. Our contributions enhance technology accessibility and robustness for computer vision applications, making it an affordable industry solution.

\bibliographystyle{ieeetr}
\bibliography{main}

\end{document}